*Article*

# Exploring the Use of Contrastive Language-Image Pre-Training for Human Posture Classification: Insights from Yoga Pose Analysis

Andrzej D. Dobrzycki [1], Ana M. Bernardos [1,*], Luca Bergesio [1], Andrzej Pomirski [2] and Daniel Sáez-Trigueros [3]

1 Information Processing and Telecommunications Center, ETSI Telecomunicación, Universidad Politécnica de Madrid, Av. Complutense, 30, 28040 Madrid, Spain; daniel.dobrzycki@alumnos.upm.es (A.D.D.); luca.bergesio@upm.es (L.B.)
2 Alexa AI, Aleja Grunwaldzka 472, 80-309 Gdańsk, Poland; pomirsa@amazon.com
3 Alexa AI, C. de Ramírez de Prado, 5, 28045 Madrid, Spain; dsaez@amazon.co.uk
* Correspondence: anamaria.bernardos@upm.es

**Abstract:** Accurate human posture classification in images and videos is crucial for automated applications across various fields, including work safety, physical rehabilitation, sports training, or daily assisted living. Recently, multimodal learning methods, such as Contrastive Language-Image Pretraining (CLIP), have advanced significantly in jointly understanding images and text. This study aims to assess the effectiveness of CLIP in classifying human postures, focusing on its application in yoga. Despite the initial limitations of the zero-shot approach, applying transfer learning on 15,301 images (real and synthetic) with 82 classes has shown promising results. The article describes the full procedure for fine-tuning, including the choice for image description syntax, models and hyperparameters adjustment. The fine-tuned model, tested on 3826 images, achieves an accuracy of over 85%, surpassing the current state-of-the-art of previous works on the same dataset by approximately 6%, its training time being 3.5 times lower than what is needed to fine-tune a YOLOv8-based model. For more application-oriented scenarios, with smaller datasets of six postures each, containing 1301 and 401 training images, the fine-tuned models attain an accuracy of 98.8% and 99.1%, respectively. Furthermore, our experiments indicate that training with as few as 20 images per pose can yield around 90% accuracy in a six-class dataset. This study demonstrates that this multimodal technique can be effectively used for yoga pose classification, and possibly for human posture classification, in general. Additionally, CLIP inference time (around 7 ms) supports that the model can be integrated into automated systems for posture evaluation, e.g., for developing a real-time personal yoga assistant for performance assessment.

## 1. Introduction

The accurate classification of human postures is key for a wide range of applications in different fields, from ergonomics to physical rehabilitation or sports. This process may require subjective and costly interpretation in terms of time and human resources and usually involves visual analysis of images or videos [1,2], digital human modelling [3], or the incorporation of virtual reality to analyze various postures in a controlled digital environment in real time [4]. In this context, the use of computer vision techniques to automate the detection and classification of postures is being widely used, often in combination with different types of "non-visual" sensors to improve the detection accuracy (e.g., motion sensors, gyroscopes, accelerometers, depth sensors, and pressure sensors).

In this article, the use of Contrastive Language-Image Pretraining (CLIP) is proposed for human posture classification (in particular, for yoga poses). CLIP is a state-of-the-art multimodal learning technique that jointly understands images and text. The model is trained on a large and diverse corpus of image–text pairs, enabling it to perform zero-shot or few-shot learning on certain tasks. Within this research, it is shown that CLIP can be effectively adapted to (yoga) pose classification by using appropriate text descriptions for each pose and applying fine-tuning on different subsets of the Yoga-82 dataset. With the proposed workflow, CLIP outperforms the current state-of-the-art methods on yoga pose classification, achieving over 85% accuracy on 82 classes. This article also provide insights into the model's behavior and limitations and discusses the potential applications of CLIP for yoga pose analysis.

The CLIP model, introduced by OpenAI in 2021, has shown significant progress in the joint understanding of images and text [5]. CLIP is a multimodal technique tailored for zero-shot learning, an approach aiming to create a model capable of effectively associating accurate image–text pairs without requiring explicit training [5]. The final objective is to correctly predict associations between images and corresponding descriptions. CLIP simultaneously trains both an image encoder (a vision model) and a text encoder (a transformer) to produce embeddings (vectors) for the image–text pairs and to then represent them in the same embedding space and perform an optimization process that maximizes the cosine similarity between the positive pairs (an image and its correct description) and that minimizes the rest (an image and text not correctly describing it).

By following this approach, novel images can be introduced to the trained model alongside a set of potential descriptions. The system then yields the most suitable text description for the given image. Consequently, CLIP's central concept revolves around acquiring visual and textual representations that hold applicability across a diverse variety of tasks, without necessitating fine-tuning. However, it is worth noting that OpenAI has not revealed the specific training data for their models. This confidentiality may lead to potential challenges in addressing domain-specific issues, which could result in reduced performance.

In this context, this article addresses the classification of human postures, more specifically, yoga poses, by exploring the use of CLIP as a classification technique through a comprehensive transfer learning procedure including data preparation, image description syntax choice, hyperparameters, and model selection. The specific application domain, yoga, is challenging in terms of body shapes, clothing, lighting conditions, and individual-dependent variations in pose execution. The classification mechanism aims to be part of a system that can automatically recognize yoga poses from visual data, ideally enabling real-time feedback and guidance for practitioners in a real-world application. This system should be easy to scale up in terms of the number of postures and in terms of application domain, so the zero/few-shot approaches for training are relevant. Thus, the main contributions in this article include a methodology regarding the use of CLIP for posture classification (including text description syntax for images and evaluation on the training dataset image quality), an exhaustive assessment of CLIP performance for a problem with a large number of postural classes and of the training frugality for the model, and a comparative performance analysis of the specific classification task between CLIP and a representative state-of-the-art algorithm used for object detection and classification (YOLOv8).

The article is structured as follows. Section 2 compiles a review on the state-of-the-art of artificial vision-based techniques for pose analysis and classification, while summarizing the existing previous (few) works using CLIP technique as a classifier. Section 3 describes the methodology for performing the fine-tuning to build the CLIP-based classifier. Section 4 gathers performance results, showing its potential on small and large multiclass datasets while also looking to determine how to build a scalable training strategy to include new poses. Section 5 compares CLIP against other state-of-the-art algorithms of the You Only Look Once (YOLO) family, while Section 6 discusses the obtained results, especially in terms of accuracy and fine-tuning and inference time. Finally, Section 7 concludes the article with further steps for research.

## 2. Related Work

### 2.1. Vision-Based Approaches for Posture Classification

Posture classification is the task of identifying the body position of one or more individuals using techniques such as computer vision or machine learning algorithms. From the early 2000s, sensor-based approaches have been proposed for recognizing human posture [6], relying on pressure mats, accelerometers, gyroscopes, magnetic trackers, etc. Sensors can be used to capture the information about the position and orientation of some part of the body. They are generally small, accurate, and robust to external conditions. Due to these features, they are still used for some specific applications and are preferred to cameras [7,8]. However, they also have some drawbacks, such as being more expensive, more cumbersome, more prone to noise and drift, and less scalable to multiple individuals.

Cameras can directly capture the visual information of body movements and poses in a non-invasive way, providing high resolution and real-time data. For these reasons, they are widely used in sports and other physical activities [9–11], surveillance [12,13], healthcare [14,15], etc. The authors of [16] used a Kinect v2 sensor to acquire both RGB and depth images of the entire body in standing, bending, sitting, walking, and crouching postures. They compared the classification results of Convolutional Neural Network (CNN) and Support Vector Machine (SVM) approaches: the CNN had a slightly better performance, but both systems reached an accuracy of 90%. Another common method of classification is the rule-based approach [17,18]. In this case, the body joints (e.g., those detected by a Kinect v2 sensor) are processed by an expert system that uses a set of rules defined using the body joint relative positions to classify a posture. This method also provides good results in movement detection, reaching a 98.5% accuracy for simple movements such as rotating an arm, clapping, or waving [17]. These approaches that take advantage of joints' position analysis are called skeleton-based [19].

Other researchers directly rely on pure RGB camera(s) and do not focus on the joints, adopting a non-skeleton-based strategy. For example, they might apply a CNN to directly process the images. This is the case in [20], where the authors only used an RGB camera and a CNN to estimate and correct the pose of fitness exercises. In this case, two ML algorithms worked together: one to classify the posture and the other to evaluate its correctness. In this work, the authors considered four dumbbell exercises: Bicep Curl, Front Raise, Shoulder Shrug, and Shoulder Press, reaching accuracies of of 89%, 100%, 89%, and 76%, respectively. The dataset used for the training was composed of 100 home-made videos created by the authors themselves with the help of a personal trainer to manually determine the correctness of the exercises. In [21], a similar system composed of an RGB camera and CNN was proposed to recognize the posture of workers. The authors divided the body into three parts: arms, back, and legs. A different classifier was trained for each part, and the result was assembled in a multi-stage CNN. With this approach, they achieved an accuracy of about 94%, training the models using the Human 3.6M dataset and validating it on an ad hoc dataset populated with images of construction workers. The three classifiers only worked with three body positions for each of them (e.g., the legs classifier recognized standing, knees bent, and squatting).

Regarding datasets, it is possible to find open datasets both for general postures and sports-oriented ones. Some examples are MPI-INF-3DHP [22], Penn Action [23], NTU RGB+D [24], and Yoga-82 [25].

### 2.2. Pose Classification in Yoga

Yoga poses involve unique postures that may be performed with different levels of difficulty, styles, and variations. In the specific case of yoga pose classification, it is possible to find several works that apply some of the strategies previously mentioned. In [26], a DCNN was used to classify yoga poses, but the images were acquired by three low-resolution infrared cameras, seeking to provide a high level of privacy by using low-cost IoT devices. With an ad hoc dataset of 187,200 low-resolution images created with the help of 18 volunteers doing 26 yoga poses, they achieved a 99.8% accuracy. In [27], the authors used images acquired by an RGB camera and an ad hoc CNN, created using Pose-Net and Mobile-Net SSD models, to recognize the postures. It reached an accuracy of 99.88%, although the dataset used for training and validation only contains 7 yoga positions. The dataset has a total of 500 videos, 350 from an open-source yoga-pose dataset and 150 recorded by the authors and added to the dataset. Other works have relied on the OpenPose framework [28], a system that applies a CNN-based model to recognize the joints. In [29], this framework was applied to create a coaching system for yoga. OpenPose does not need training or fine-tuning, so the authors added a correction algorithm module, based on rules, to increase the detection accuracy. They achieved a 91% accuracy on a homemade dataset, comprising 8 yoga positions, which was created in collaboration with a professional coach, though the exact number of videos in the dataset was unspecified. In [30], the authors sought unify classification and posture evaluation by using skeleton keypoints. This study reported a classification accuracy of 83.21% over 45 postures in a dataset containing 1931 images, by following a method that implies the construction of contrastive examples. Both coarse and fine triplet examples were used to learn discriminative features from human skeleton keypoints (the coarse triplet consisted of an anchor, a positive example from the same category, and a negative example from a different category, while the fine triplet consisted of an anchor, a positive example, and a negative example from the same category with different pose qualities). Mediapipe was then used to detect 33 human body keypoints from the input pose image. Their coordinates were then fed into a pose feature encoder. Finally, pose feature encoding was performed using contrastive skeleton feature representations, using a weight-shared Neural Network (NN) to encode the skeleton keypoints into feature vectors, and a triplet contrastive loss was applied to maximize the agreement between similar pose images and minimize the agreement between dissimilar pose images in the latent feature space. A summary of the main classification works, discussed in this and the previous section, and their results, is presented in Table 1.

**Table 1.** Comparison of camera-based related works.

| Work | Goal of the Work | Datasets | Algorithms | Results (Accuracy) |
|------|------------------|----------|------------|--------------------|
| [16] | Comparing the performance of classical and ML techniques in posture recognition | Ad hoc (1040 images) | SVM vs. CNN | SVM: 93.1% CNN: 95.7% |
| [17] | Developing a pose classifier using a rule-based approach | Ad hoc (1600 videos) | Rule-based reasoning applied to body joints | 91.3% |
| [18] | Enhancing human posture recognition using a rule learning algorithm | MSR-Action3D, Microsoft MSRC-12, UTKinect-Action, Ad hoc (Baduanjin) | Feature extraction + RIPPER[31] + classification | MSR3D: 94.5% MSRC: 97.6% UTKA: 98.1% Badua.: 99.6% |
| [20] | Detecting and correcting exercise posture | Ad hoc (100 videos) | 2 CNNs | 88.5% |
| [21] | Recognizing postures of workers to improve their safety and health | Human 3.6M | multi-stage CNN | 94.8% |
| [26] | Recognizing yoga postures using IR cameras to preserve privacy | Ad hoc (187,200 images, 26 yoga poses) | DCNN | 99.8% |
| [27] | Correcting yoga poses | Yoga-pose dataset (350 videos of 7 poses) + ad hoc (150 videos) | CNN | 99.88% |
| [29] | Creating a yoga coach | Ad hoc (8 yoga poses, no more details) | OpenPose (CNN) | 91% |

The Yoga-82 dataset is specifically tailored for yoga pose classification. Specifically, the Yoga-82 dataset has been used in previous research, but insights from the literature show that, in most previous attempts, classification was performed on a much smaller set than the 82 postures available. The accuracy of some of the architectures and models found is summarized below.

- YoNet [32] is able to classify five types of yoga poses with an accuracy of 94.91%. YoNet is a deep learning model that utilizes depthwise separable convolution—a technique presented by Xception [33]—for parameter and computational reduction. To extract spatial characteristics, such as edges, YoNet utilizes a standard convolution layer on the input image, followed by multiple depthwise separable convolution layers to extract depth attributes such as the position and orientation of the human body. These two feature types are combined prior to the classification layer. For classification, YoNet utilizes two dense layers with ReLU activation and batch normalization. This enables the input image to be classified into one of five yoga poses. The ultimate layer applies softmax activation to display the probability of each class.

- Y_PN-MSSD [27] has an accuracy rate of 99.88% for a total of seven poses. The Y_PN-MSSD model architecture is a deep learning-based model that combines two components, namely PoseNet [34] and MobileNet SSD [35]. The main steps in the model are as follows:

  Feature Extraction: The PoseNet model utilizes CNNs to identify the essential human body keypoints in every picture. These points are linked to create a skeletal representation of the pose.

  Posture Recognition: The system utilizes MobileNet SSD, a Single Shot Detector (SSD), to classify the pose according to the skeletal features. The system also executes human detection and generates bounding boxes in every frame.

- The original paper presenting the Yoga-82 dataset reported a maximum accuracy of 79.35% for its Architectural Variant 1 classifying all 82 postures. Using only the Yoga-82 dataset and classifying all 82 postures, this accuracy represents the current state of the art. Architectural Variant 1 is a modification of the DenseNet-201 [36] architecture that introduces a hierarchical structure to improve classification performance by utilizing the hierarchy inherent in the dataset. This variant integrates hierarchical connections after DenseBlock 2 and DenseBlock 3, catering to Class Level 1 (with 6 classes) and Class Level 2 (with 20 classes), respectively. The fundamental idea behind this variant is that broader classes are classified in the intermediate layers of the network, while more specific classes are processed by the end layers. The initial-to-mid layers specialize in first-level classification and pass on image details to subsequent layers for second-level classification. The common layers up to DenseBlock 2 focus on capturing the fundamental pose structure, with subsequent layers further refining the model's understanding.

Almost all of the previous works gathered in this review that recognize fewer than 10 poses achieve a precision of over 90%. In general, all high-accuracy results mentioned in previous research are due not only to a well-designed architecture, but also to the availability of large training datasets for a small number of postures. As expected, a large number of postures reduces the accuracy of pose classification.

### 2.3. The Use of CLIP as a Classifier

The concept behind Contrastive Language-Image Pretraining (CLIP) involves an image label predictor that takes a distinctive approach. It undertakes the simultaneous training of an image encoder and a text encoder with the objective of correctly associating pairs from a batch of training instances, comprising both images and texts [5]. Unlike alternative methodologies that train an image feature extractor to forecast labels, CLIP prioritizes the interplay between text and image. This training procedure equips the encoders to subsequently execute zero-shot classification tasks through embedding the intended image and an assortment of conceivable labels.

CLIP primarily centers its efforts on accomplishing visual classification tasks through the straightforward presentation of feasible categories for input images. It is important to note that the model's design does not accommodate the identification of multiple classes within a single image, and it functions optimally when images exhibit minimal or no background noise. Moreover, the training regimen of CLIP involved a dataset of 400 million samples sourced from the internet, encompassing general domains such as animals and everyday objects[5].

The extent of available literature applying CLIP to real-world problems remains relatively limited. To our current knowledge, no documented instances of prior applications to posture classification exist. However, there have been instances of CLIP's utilization, such as its deployment in diagnosing radiological images[37]. In such endeavors, authors have undertaken enhancements to CLIP's pre-training process, adapting the model by incorporating new datasets encompassing unpaired images and textual data. This adaptation includes the implementation of an updated loss function. In contrast to the conventional cosine similarity approach between images and text, a soft semantic matching loss has been employed. This new loss function integrates human domain knowledge to mitigate the occurrence of false negatives.

Within the context of clinical reports derived from radiology images, one study[38] has integrated two CLIP-based models. These models are engineered to predict either individual report sentences or a synthesis of report sentences. This entails the deployment of a single image encoder and dual text encoders, contingent upon the input type. Model evaluation involves subjecting resultant reports to the original dataset labeler. This evaluation methodology computes predictions against the original labels, ultimately yielding an F1 score of 0.310 in comparison to the 0.294 obtained by focusing solely on individual reports.

For vehicle identification[39], a distinct approach has been taken, utilizing natural language queries. Unlike other problem domains, each entry comprises three instances: three frames of the same object and three corresponding queries describing said frames. Collating information from all frames and queries, an averaged feature matrix is generated, encapsulating the essence of the three frames. A separate matrix combines information from the three queries. Fine-tuning is executed via the feature extractor, and evaluation hinges on the cosine similarity metric. Performance is assessed using the Mean Reciprocal Rank (MRR), a metric suitable for systems that furnish a ranked list of responses to queries. This approach achieves a top-8 ranking in the Nvidia AI City Challenge.

In [40], a comparison is drawn between this technique and more conventional algorithms. The authors subjected CLIP and YOLO to rigorous testing for real-time video fire and gun detection, a context crucial for security applications. Despite CLIP not being inherently tailored for object detection, it remarkably surpassed YOLO's performance in both tasks. Notably, it achieved an impressive F-measure of 99.7% in contrast to YOLO's 85% for the former, and 97.3% compared to YOLO's 96.5% for the latter.

The distinctive advantage of CLIP lies in its ability to excel in these tasks without requiring the substantial labeling efforts typical of conventional object detection algorithms. This attribute facilitates swift technology adaptation, allowing for the efficient definition of relevant semantic contexts of interest.

## 3. Procedure for Setting Up CLIP as a Posture Classifier

The methodology followed for CLIP-based posture classification is rooted in the overarching goal of enhancing both the accuracy and efficiency of this task while concurrently minimizing the training requirements. This approach is driven by the aspiration to create a scalable system capable of accommodating a wide spectrum of postures. The methodology encompasses several key components, each meticulously designed to contribute to the successful setup of CLIP as a posture classifier. These components include a rigorous analysis of the dataset used for training and evaluation, the configuration of the CLIP model, and the definition of hyperparameters. Within the configuration of CLIP, there are two pivotal subcomponents: the syntax for image description and the baseline zero-shot evaluation, as well as the careful selection of the visual encoder and the definition of hyperparameters for the fine-tuning process. The procedure aims at supporting the use of CLIP for posture classification, ensuring that the model is capable of accurately and efficiently recognizing a wide array of postures while maintaining scalability and adaptability. The following sections will explain each step in detail.

### 3.1. Dataset Analysis

The Yoga-82 dataset has been chosen for the analysis presented in Sections 2.1 and 2.3; this dataset contains samples for 82 different yoga poses. Yoga-82 includes not only the basic yoga postures, but also their variations, representing different levels of difficulty or adaptations of the core postures. It includes images in-the-wild, considering a variety of viewpoints, lighting conditions, resolutions, and occlusions, an aspect that can be considered as positive in order to guarantee a robust classifier performance, but that makes the classification process more challenging. Moreover, it also contains synthetically generated images (e.g., creative drawings).

Originally, the dataset contained over 28.4 K images of yoga poses spread across the target 82 classes. However, as the dataset retrieves images based on their URLs, a total of 19.1 K images could be downloaded for analysis in Spring 2023. The dataset contains a three-level hierarchy that includes body positions, variations of body positions, and the names of the postures. The 82 classes (Level 3, L-3) are merged/collapsed into 20 superclasses (Level 2, L-2) based on similarities in body postures, which in turn are merged and regrouped into 6 superclasses at the top level of the hierarchy (Level 1, L-1).

The dataset has a variable number of images in each class, ranging from 38 (minimum) to 788 (maximum), with an average of 233 images per class. In each image, there is one or more people performing the same yoga pose. In addition, the images also contain poses that were taken from different points of view of the camera. The background in which the yoga poses are being carried out is highly varied, ranging from clean to natural or human-built backgrounds (e.g., forests, beaches, interiors, etc.). Some images are synthetic, containing only silhouettes, sketches, and drawings of yoga poses. For our analysis, these images were kept. The distribution of the L-1 and L-2 superclasses is illustrated in Figure 1. The unbalanced nature of each of the classes at the L-3 level of the hierarchy (82 classes), as well as their relationship to the L-2 level, is shown at Appendix A.

From the dataset retrieved (19.1 K images), two subsets have been extracted, in order to perform the first CLIP assessment. Each subset contains the same six postures (with Balasana, Dhanurasana, Marjaryasana, Sarvangasana, Ustrasana, and Utkatasana postures), which have been selected because they cover a wide range of physical positions, including sitting and standing. These poses vary in physical execution, exposing the model to different body orientations, postural complexities, and limb positions. From beginner-friendly to more advanced, the selected poses also offer different levels of difficulty. This variety may permit to test the model's versatility in its capacity to recognize and adjust to users with differing skills. Each of the six postures targets specific body regions and muscle groups. For example, Marjaryasana mainly activates the muscles of the back and spine, while Ustrasana necessitates deep backbending, requiring flexibility in the anterior of the body. This variety aims at assess the efficiency of the model to identify and classify postures regardless of the specific muscles and body areas they emphasize.

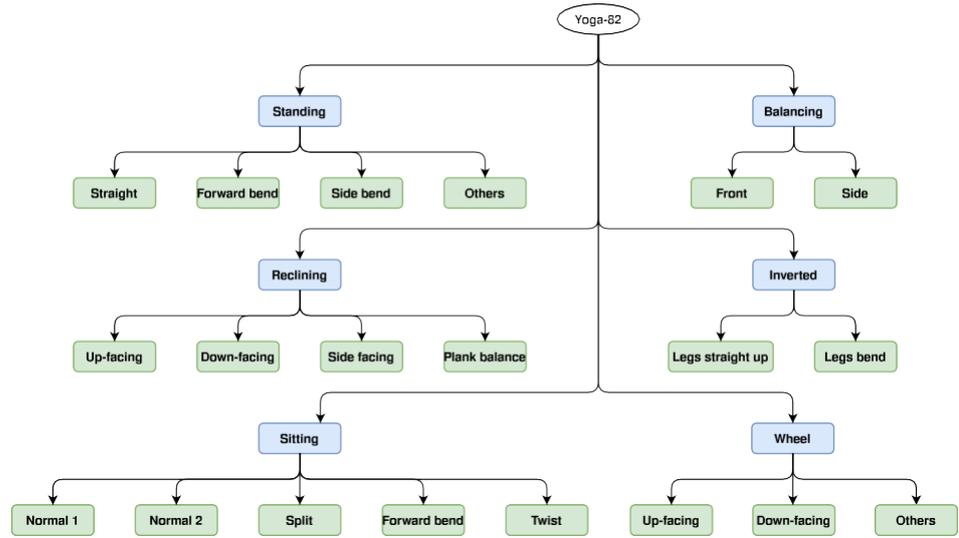

**Figure 1.** Distribution of L-1 (blue) and L-2 (green) superclasses in the Yoga-82 dataset.

A manual filter was applied to one of the subsets, in which only the images with the highest quality and on clean backgrounds were retained, leaving a total of 543 images (Yoga-82-Subset II). No manipulation has been done to the other subset of 1627 images (Yoga-82-Subset I). Figure 2 shows an example image of each of the subsets created. Tables 2 and 3 show the number of samples for each of the six classes, both for Subset I and II.

A curated third validation subset was created to compare CLIP and YOLOv8 in classifying the 82 postures fairly, and the specific use of this subset is described in Sections 4.3 and 5 with the images available at [41]. After eliminating the images that were also present in the CLIP training set, a total of 3487 images remained in the validation subset (YoPo-Subset).

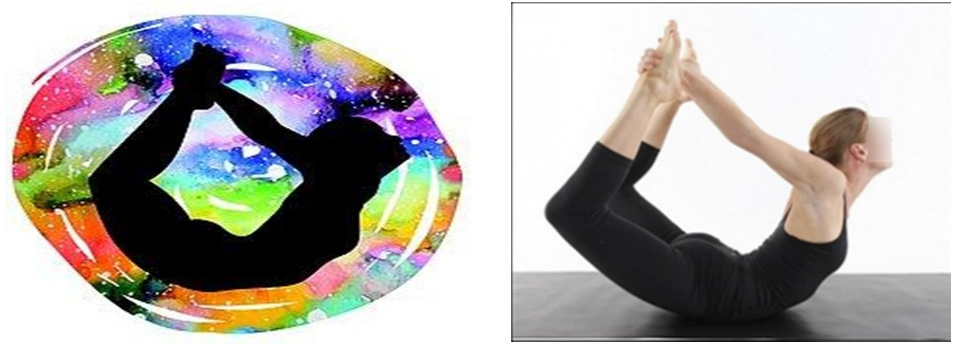

**Figure 2.** Dhanurasana pose of Subset I (**left**) and Dhanurasana pose of Subset II (**right**).

**Table 2.** Number of images per class for unfiltered subset (Yoga-82-Subset I).

| Unfiltered Subset (Yoga-82-Subset I) | | |
|---|---|---|
| | **Train** | **Test** |
| Dhanurasana | 171 | 45 |
| Ustrasana | 226 | 51 |
| Sarvangasana | 192 | 51 |
| Utkatasana | 215 | 58 |
| Marjaryasana | 291 | 61 |
| Balasana | 206 | 60 |

**Table 3.** Number of images per class for filtered subset (Yoga-82-Subset II).

| Filtered Subset (Yoga-82-Subset II) | | |
|---|---|---|
| | **Train** | **Test** |
| Dhanurasana | 47 | 10 |
| Ustrasana | 61 | 11 |
| Sarvangasana | 43 | 15 |
| Utkatasana | 98 | 21 |
| Marjaryasana | 114 | 21 |
| Balasana | 71 | 31 |

### 3.2. CLIP Configuration

The process of fine-tuning the CLIP model for yoga pose classification began with an analysis of the Yoga-82 dataset. Subsets of this dataset, specifically Yoga-82-Subset II, were prepared based on the yoga postures. An initial evaluation was performed on this filtered subset to test the model's zero-shot performance. The syntax structure for the image descriptions (prompts) and the visual encoder were then chosen as part of the model's configuration. Subsequently, multiple rounds of fine-tuning were conducted using Yoga-82-Subset I and II. This step was crucial to assess the influence of image quality and to determine the optimal hyperparameters for the model. Upon obtaining satisfactory results from these steps, the CLIP model was fine-tuned for all 82 postures present in the Yoga-82 dataset. The performance of the model was then evaluated as detailed in Section 4 of this paper. The model's training efficiency was also assessed, and its performance was benchmarked against the YOLO model. A flowchart illustrating the process is provided in Figure 3.

All experiments were performed using a development environment that includes a computer with an Intel(R) Core(TM) i9-12900K CPU with 16 cores and 2 threads per core and that operates at a maximum frequency of 5.2 GHz, and it has 2 cache levels (L1 and L2) for improved data access and 32 GB of RAM. It also has a NVIDIA RTX A4000 graphic card, enabling parallel processing on GPUs for faster computation. The specific driver version is 535.104.05, which supports CUDA 12.2.

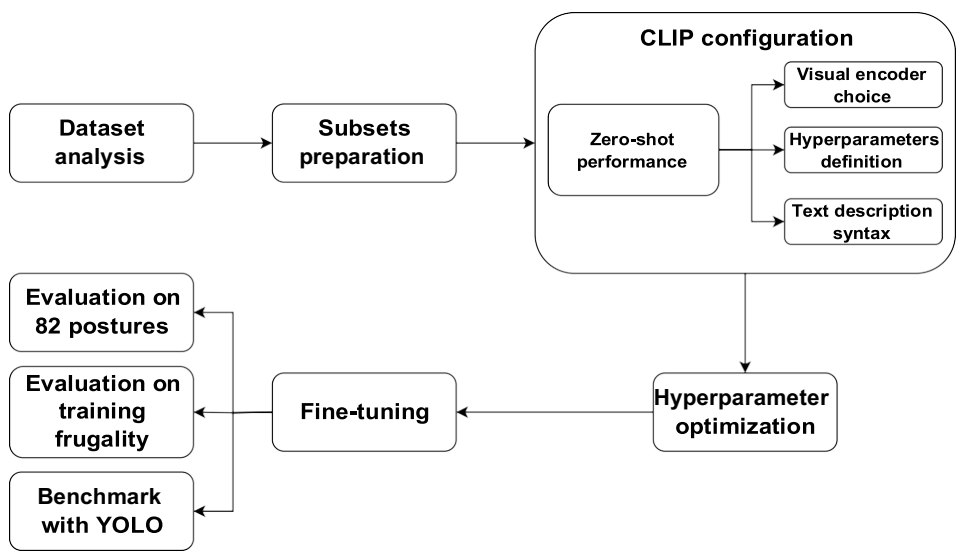

**Figure 3.** Flowchart of the procedure followed to set up CLIP as a posture classifier.

### 3.2.1. Syntax for Image Description and Baseline Zero-Shot Evaluation

CLIP acts as a framework for aligning image–text pairs and uses contrastive learning to establish contextual similarity relationships between them. In this context, the image–text alignment problem involves establishing the association between images and captions.

The primary goal of contrastive learning is to differentiate between matched image–text pairs (referred to as "positives") and unmatched pairs (referred to as "negatives") [42,43]. During inference, hand-engineered text prompts are used, e.g., "a photo of a <category>", as a query for the text encoder, e.g., a transformer [44]. The output text embeddings are matched with the visual embeddings from an image (visual) encoder, e.g., ResNet [45] and ViT [46], to predict the output class. Designing high quality contextual prompts have been proven to enhance the performance of CLIP and other Vision-Language models [47,48]. In this experiment, there were comparable results for various image–text pairs by altering the prompts utilized for zero-shot classification.

Table 4 shows the accuracy for each of the pre-trained clip models performing zero-shot classification on the filtered subset (Subset II), i.e., without fine-tuning the visual encoder models. The text descriptions utilized to annotate the images followed this format: "Yoga pose <category>". The algorithm's accuracy was assessed by initially dividing the dataset into training and testing/validation sets using an 80–20 split. The accuracy of the algorithm was then calculated by determining the ratio of correct predictions to the total number of predictions in the designated test set. Results show that the classifier's performance is similar to random behaviour.

**Table 4.** Zero-shot performance over filtered subset (Subset II).

| Visual Encoder Model | Accuracy |
|---|---|
| | Top-1 |
| RN50 | 20.8% |
| RN101 | 41.5% |
| RN50x4 | 30.2% |
| RN50x16 | 49.1% |
| RN50x4 | 37.7% |
| ViT-B/32 | 24.5% |
| ViT-B/16 | 30.2% |
| ViT-L/14 | 35.8% |
| ViT-L/14@336 px | 28.3% |

### 3.2.2. Visual Encoder Choice and Hyperparameters Definition for Fine-Tuning

CLIP's visual encoder diverges from conventional visual encoders in Visual and Language (V&L) models. Whereas conventional encoders rely on region-based, such as a BUTD [49] object detector, or grid-based approaches [50] and are pre-trained on annotated visual datasets, CLIP uses a distinct technique. It derives visual representations by monitoring natural language. CLIP implements a "shallow interaction design", where a visual encoder and a text encoder independently encode input images and text. In the development of CLIP, two distinct architectural approaches were considered for the image encoder. The first architecture entails the utilization of ResNet-50, a widely adopted and well-established base architecture renowned for its commendable performance characteristics [45]. This selection was improved through various modifications, incorporating enhancements obtained from ResNet-D [51] among other sources. Furthermore, the conventional global average pooling layer was substituted with an attention pooling mechanism. The attention pooling is realized as a singular layer, designed in a "transformer-style" framework featuring multi-head QKV attention, where the query is conditioned upon the globally averaged representation of the image.

In contrast, the second architectural exploration entailed an investigation into the Vision Transformer (ViT) [46]. In this instance, the implementation closely adhered to the original ViT model, with the sole modification being the inclusion of an additional layer normalization applied to the amalgamated patch and position embeddings preceding the transformer. Furthermore, a slightly distinct initialization scheme was employed.

The culmination of the architectural exploration led to the training of a series of five ResNets and three Vision Transformers. Within the ResNet category, the models encom-passed ResNet-50, ResNet-101, and three additional variants following an EfficientNet-style model scaling approach, each demanding approximately 4 times, 16 times, and 64 times the computational resources of a ResNet-50. These variants were denoted as RN50x4, RN50x16, and RN50x64, respectively. Concurrently, in the Vision Transformer domain, models included ViT-B/32, ViT-B/16, and ViT-L/14.

The choice for all the experiments was the ViT-B/32 architecture. The decision to utilize this architecture in the CLIP framework stemmed from careful consideration of various factors such as input size and hardware limitations during fine-tuning and inference. This decision was based on the need to balance performance and computational efficiency, particularly in situations with limited resources. The ViT-B/32 architecture applies pre-processing steps to the images, including resizing them to a resolution of 224 × 224 pixels.

In order to adapt the CLIP model to the specific domain of posture image classification, the fine-tuning and optimization of the hyperparameters play a crucial role. Before starting to classify the 82 postures of the Yoga-82 dataset, an analysis on the Yoga-82-Subset I and II (6 classes) was conducted to determine whether there were significant differences in the accuracy obtained due to the cleanliness of the training images. The conclusion of this analysis are elaborated further in Section 4.1.

In the fine-tuning stage, the decision was to associate the text description "Image of a person doing the yoga pose <category>", given the similarity of results observed in the zero-shot approach and the outcomes of Table 5, with various fine-tuning attempts made with the filtered subset by changing the description syntax. Even though the results indicate that the "Yoga pose <category>" description yields better accuracy, the difference of less than 1% compared to the chosen description and the timing of the later experiment led to the decision to not adopt the improved description and redo the experiments. A difference of less than 4% in accuracy was observed among all tested syntax. Specifically, the mentioned syntax achieved an accuracy of 99.1%. The increase in accuracy when translating Sanskrit classes into numbers during classification is remarkable and may be attributed to the limited number of Sanskrit words found in the CLIP pre-training set's 400 million image–text pairs.

For the fine-tuning, we followed the method described in the original CLIP paper [5]. In particular, we used the loss function described in that work:

$$Loss = \frac{Loss_{img} + Loss_{txt}}{2} \tag{1}$$

where both the loss function of the images and of the text are calculated using the cross entropy loss. These functions are used to measure the distance between the logits $L_i$ and the truth values $T_i$.

$$Loss_x = -\sum_i^C T_i \log L_i \tag{2}$$

Because the dataset contained a large number of images and labels, the fine-tuning process was divided into many batches. The total loss function for each epoch was calculated by summing the loss of each batch and then averaging it at the end of the epoch.

After selecting the description associated with each class, we proceeded to adjust the hyperparameters of the model architecture. The chosen visual encoder was the Vision Transformer (ViT) ViT-B/32 with 224 × 224 input resolution [5]. By tuning these hyperparameters, the aim was to find the optimal configuration that maximizes the performance of the model in the desired task. The main hyperparameters were the learning rate and the weight decay. Our decision to focus on exploring the hyperparameters of the learning rate and weight decay during the fine-tuning process was motivated by hardware constraints that prevented us from conducting a comprehensive grid search. These limitations required us to emphasize hyperparameters with significant effects. The learning rate was critical, and initiating with a conservative global learning rate, e.g., 10 times lower than that used during the pre-training phase, fostered stable convergence throughout the fine-tuning process. Weight decay served as a crucial regularization technique that combated overfitting, particularly in scenarios with large models, such as this one, or small datasets.

The focus on these two hyperparameters enabled a balance between computational efficiency and the optimization of fine-tuning, further enabling us to examine the essential factors influencing training dynamics and generalization performance.

**Table 5.** Relationship between accuracy and description syntax selected after fine-tuning using Subset II.

| Text Description Syntax | Accuracy |
| --- | --- |
| "Image of a person doing the yoga pose <category>" | 99.1% |
| "Yoga pose <category>" | 100% |
| "<category>" | 96.3% |
| "<category>" to numeric | 98.2% |

The common CLIP hyperparameters values [5] were used as a starting point, but reduced by 1–2 orders of magnitude. After several tests, the best accuracy results on both six-class subsets were obtained for a learning rate of $10^{-5}$, a weight decay of $10^{-3}$, and 5 epochs.

The conclusion of this experiment is that filtering the full dataset of 19.1K images and 82 classes is not needed to achieve a fine-tuned working CLIP model, as the task is costly and the accuracy of the model when fitting its parameters to the filtered and unfiltered subsets varies by about 1%. Apart from this, the existence of imperfect image samples may increase the robustness of the classifier in terms of applicability to real-life conditions. Figure 4 shows the difference between the image–text pairs' cosine similarity before and after fitting the model parameters with the filtered subset. It is observable that, before fine-tuning, all cosine similarities were high, possibly due to the presence of the word "person". After fine-tuning, the yellow squares accurately indicated where the text and image match. While maximizing cosine similarity for the yellow squares on the main diagonal, the goal was to simultaneously minimize similarity when the image–text pairs do not align.

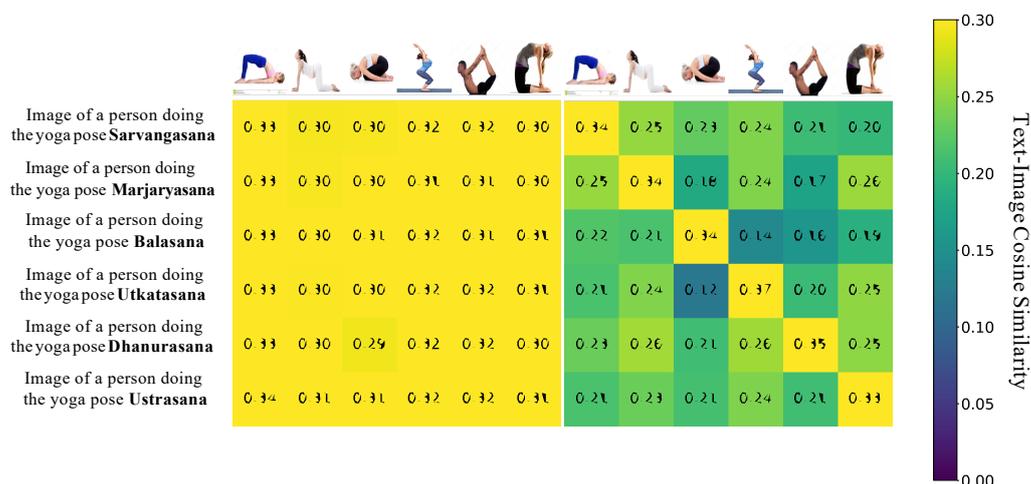

**Figure 4.** Cosine similarity between the text and image features was analyzed using a zero-shot approach (**left**) and after fine-tuning (**right**) over the filtered subset.

## 4. Evaluation of Fine-Tuned CLIP Performance

This section gathers the results obtained through a validation strategy that evaluates the performance of the model on the unfiltered six-posture (Yoga-82-Subset I), filtered six-posture (Yoga-82-Subset II), and 82-posture datasets.

A strategy organized by posture type was implemented to perform the validation. This allows an exhaustive evaluation of the model on the different hierarchical categories in which the Yoga-82 dataset is organized. The validation on the subsets of 6 classes was performed only on the third level of classification (L3).

### 4.1. Fine-Tuned CLIP Performance for Six-Posture Subsets

The first validation was performed on Yoga-82-Subset I and II. These subsets allowed us to evaluate the ability of the model to recognize and discern a small group of postures accurately. The training and testing/validation split resulted in 1301 training images and 326 test images for Subset I (1627 images) and 434 training images and 109 test images for Subset II (543 images) using an 80–20 split for each one of the subsets. When fitting the model parameters with both subsets, 5 epochs and a batch size of 6 were considered. The results of this first validation for the unfiltered subset (Subset I) are shown in Table 6, and those for the filtered subset (Subset II) are shown in Table 7. The choice of hyperparameters was found to have a profound effect on the performance of CLIP. In particular, a learning rate of $10^{-5}$ emerged as the most effective setting, yielding the highest top-1 accuracy of 0.988 for Subset I and of 0.991 for Subset II at the same learning rate. This observation underscores the importance of using a lower learning rate for fine-tuning tasks when working with CLIP, as higher values such as $5 \times 10^{-4}$ or $10^{-4}$ resulted in decreased accuracy, suggesting that overly high learning rates can hinder model convergence and degrade performance.

In addition, the weight decay hyperparameter was found to be a critical factor in influencing model performance. In particular, a weight decay of $10^{-3}$ combined with a learning rate of $10^{-5}$ produced the best results when working with both Subset I and II. This finding highlights the importance of striking a delicate balance between the learning rate and the weight decay for optimal fine-tuning results.

These experiments underscore the need for the careful selection and fine-tuning of hyperparameters to achieve peak performance, as the observed variations in accuracy underscore their significant influence on the fine-tuning process.

**Table 6.** Experiments in which CLIP is fine-tuned with Subset I.

| Model | Accuracy | Hyperparameters | |
|---|---|---|---|
| | **Top-1** | **Learning Rate** | **Weight Decay** |
| | 17.8% | $10^{-4}$ | $10^{-4}$ |
| | 18.4% | $5 \times 10^{-4}$ | $10^{-3}$ |
| | **98.8%** | **$10^{-5}$** | **$10^{-3}$** |
| ViT-B/32 | 98.2% | $10^{-5}$ | $10^{-4}$ |
| | 85.3% | $10^{-6}$ | $10^{-3}$ |
| | 77.9% | $10^{-6}$ | $10^{-4}$ |
| | 97.2% | $5 \times 10^{-6}$ | $10^{-3}$ |
| | 96.9% | $5 \times 10^{-6}$ | $10^{-4}$ |

Best results shown in bold.

**Table 7.** Experiments in which CLIP is fine-tuned with Subset II.

| Model | Accuracy | Hyperparameters | |
|---|---|---|---|
| | **Top-1** | **Learning Rate** | **Weight Decay** |
| | 31.2% | $10^{-4}$ | $10^{-4}$ |
| | 30.3% | $5 \times 10^{-4}$ | $10^{-3}$ |
| | **99.1%** | **$10^{-5}$** | **$10^{-3}$** |
| ViT-B/32 | 98.2% | $10^{-5}$ | $10^{-4}$ |
| | 72.5% | $10^{-6}$ | $10^{-3}$ |
| | 74.3% | $10^{-6}$ | $10^{-4}$ |
| | 96.3% | $5 \times 10^{-6}$ | $10^{-3}$ |
| | 97.2% | $5 \times 10^{-6}$ | $10^{-4}$ |

Best results shown in bold.

### 4.2. Fine-Tuned CLIP Performance for 82 Postures

The validation was then extended to the full dataset consisting of 82 different postures using a learning rate of $10^{-5}$ and a weight decay of $10^{-3}$, taking into account the results of fine-tuning CLIP with Subset I and II. Figure 5 shows the fine-tuning curve of the resulting CLIP posture classifier when simulating for 5 epochs using a batch size of 82. It is noteworthy that the model achieved an accuracy of over 77% in classifying the 82 classes after a single epoch. After the initial epoch, there is a noticeable decrease in the slope of the accuracy curve. Table 8 shows the results of fine-tuning the CLIP model on the Yoga-82 dataset structured into three hierarchical levels: L1, L2, and L3, comprising 6, 20, and 82 classes, respectively.

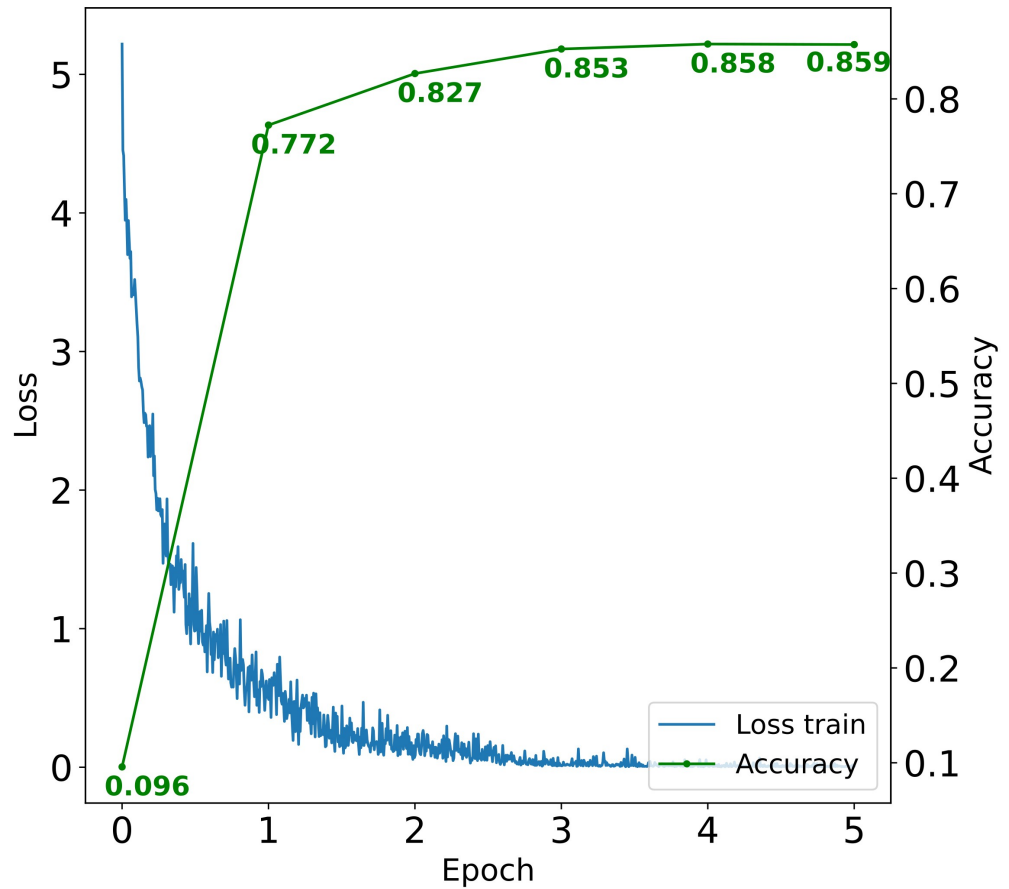

**Figure 5.** Fine-tuning curve of CLIP.

**Table 8.** Results of fine-tuning CLIP with the Yoga-82 dataset.

| Model | Accuracy (%) | | | | | | Hyperparameters | |
|---|---|---|---|---|---|---|---|---|
| | **Top-1** | | | **Top-5** | | | **Learning Rate** | **Weight Decay** |
| ViT-B/32 | L1 | L2 | L3 | L1 | L2 | L3 | | |
| | 94.2 | 90.7 | **85.9** | 99.9 | 98.8 | 96.8 | $10^{-5}$ | $10^{-3}$ |

Best results shown in bold.

Notably, in the most challenging classification scenario (L3) with 82 classes, the model achieved a commendable top-1 accuracy of 85.9%. This performance underscores the model's ability to handle complex, fine-grained classification challenges and indicates its potential to excel in real-world scenarios where class diversity is substantial. The observed decline in accuracy from L1 to L3 is consistent, as handling a greater number of classes requires enhanced discriminative capabilities. These results highlight the flexibility of the model and its potential utility across a spectrum of real-world applications requiring varying degrees of classification complexity.

Due to the unbalanced nature of the dataset, it is not feasible to perform traditional cross-validation. This would force us to reduce the number of images per class according to the most restrictive class (38 images) in order to perform the K partitions. Therefore, the dataset was randomly divided into a training set and a test set 78 times, the model was fine-tuned each time, and the accuracy results were averaged. This improved the assessment of the model's performance.

The results obtained were satisfactory, surpassing the previous state of the art in classification on the Yoga-82 dataset with an accuracy of 85.94%. The CLIP model has demonstrated a remarkable ability to accurately classify postures, even in the case of the most complex and challenging postures. Figure 6 shows the activation map obtained on different images of the dataset with a zero-shot approximation and after fine-tuning. The fine-tuned CLIP model examines the particular grip of the limb in the middle row image and the position of the arm and head in the lower row. These maps were computed using gScoreCAM [52] (k = 300) to better understand the behavior of CLIP after fine-tuning. gScoreCAM is a method for visualizing the main objects that a model, such as OpenAI's CLIP, is focusing on in an image. It works by visualizing the top 10% most sensitive (highest-gradient) channels. This method obtains state-of-the-art weakly supervised localization results using CLIP (in both ResNet and ViT versions).

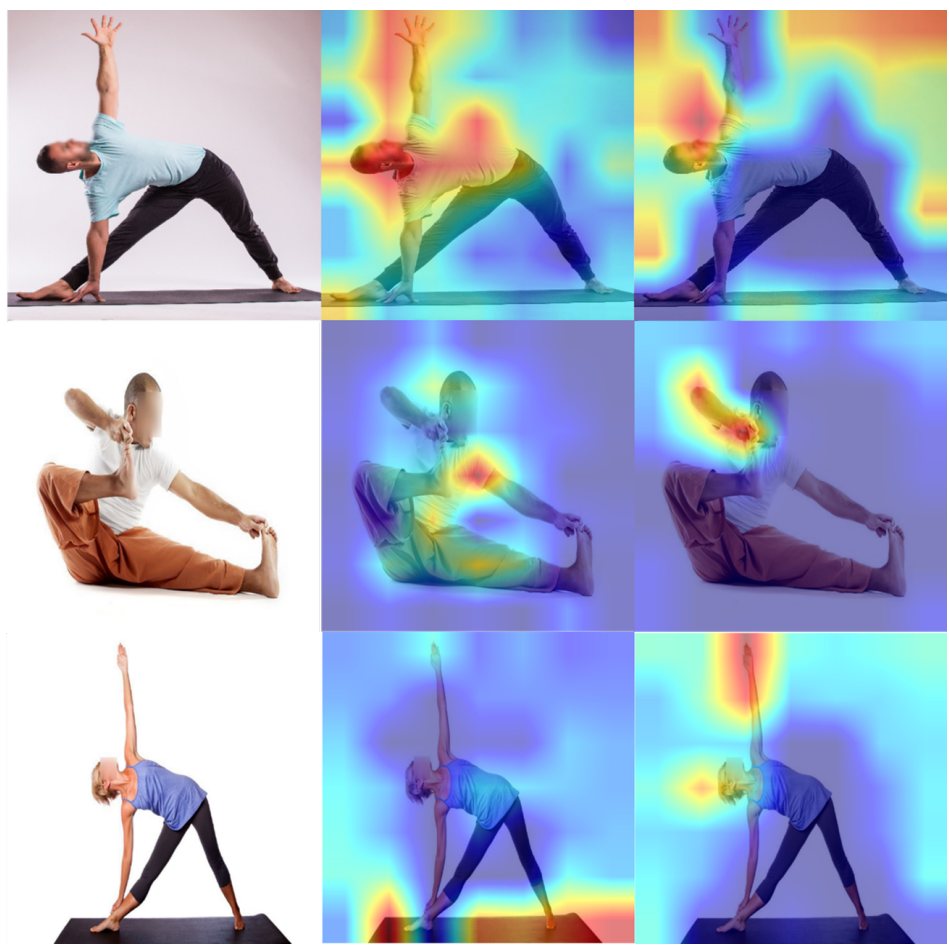

**Figure 6.** Activation map samples with CLIP. For each sample shown (**left**), corresponding activation maps of the CLIP zero-shot (middle) and after fine-tuning CLIP (**right**) are shown. The description used was "Image of a person doing the yoga pose <category>".

Multiple representations of the confusion matrix were created based on the L-1 superclass hierarchy due to the excessive size of the confusion matrix with 82 classes. Six confusion matrices were created, one for each L-1 superclass, to facilitate the identification of the yoga postures for which CLIP has the most difficulty classifying. This approach of presenting the six confusion matrices also made it possible to confirm the similarity between postures belonging to the same superclass. The confusion matrices generated for the L-1 superclasses "Sitting" and "Reclining", and normalized over the predicted (columns) values, are shown in Appendix B. For example, it can be observed that CLIP has problems classifying the postures *Makara Adho Mukha Svanasana* and *Chaturanga Dandasana* for the superclass "Reclining" even after fine-tuning. This is likely due to two factors. First, there is a relatively small sample size of 46 images of the *Makara Adho Mukha Svanasana* posture in the dataset. Secondly, the significant similarity between the two postures may also play a critical role, as depicted in Figure 7. This issue could potentially be resolved by increasing the number of images of the *Makara Adho Mukha Svanasana* pose in the dataset.

Fine-tuning CLIP for posture recognition yielded promising results, leading to the conclusion that CLIP can be fine-tuned for posture recognition. These findings support the results of [53], wherein it was also concluded that CLIP itself is a highly effective fine-tuner. Table 9 shows the accuracy obtained for the different hierarchical levels after the 78 fine-tuning iterations. Remarkably, L1, L2, and L3 yielded top-1 weighted accuracies of 93.9%, 90.1%, and 84.1%, respectively, showcasing a clear decline in accuracy with increasing complexity in classification tasks. To provide a comprehensive assessment, the analysis includes mean values and standard deviations. Notably, the standard deviation exhibits minimal variation, particularly significant at the L3 level, where the variation is less than 1%. This slight variation highlights the dependability and consistency of the model, even when dealing with complex classification tasks. Table 10 shows the top performance metrics of accuracy, precision, recall, and F1 score after adjusting CLIP 78 times to solve the classification problem with the aforementioned data. The precision, recall, and F1 score are shown as weighted averages due to the unbalanced nature of the dataset. On a test and validation set of 3826 images, an F1 score of 85.7% and an accuracy of 85.9% were obtained.

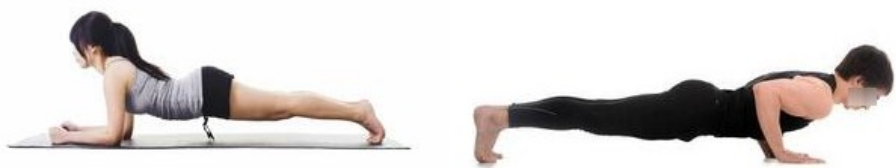

**Figure 7.** Makara Adho Mukha Svanasana (**left**) and Chaturanga Dandasana (**right**) poses extracted from the Yoga-82 dataset.

**Table 9.** Top-1 weighted accuracies.

|    | **Mean** | **Standard Deviation** |
|----|----------|------------------------|
| L1 | 93.9%    | 0.5                    |
| L2 | 90.1%    | 0.6                    |
| L3 | 84.1%    | 0.9                    |

**Table 10.** Performance of CLIP. Best metrics are reported.

|    | **Accuracy** | **Precision** | **Recall** | **F1** | **Support** |
|----|--------------|---------------|------------|--------|-------------|
| L3 | 85.9%        | 86.1%         | 85.9%      | 85.7%  | 3826        |

### 4.3. Notes on Computational Cost

The fine-tuning time for the 5 epochs took 13.8 min on the system described in Section 3.2.

The inference time, calculated as an average over the 3487 images of the validation set (YoPo-Subset), was 7.1 ms. This time includes image preprocessing, tokenization of the sequences of the given text input(s), encoding by the vision and text portions of the CLIP model, and the computation of logit scores corresponding to each image–text pair.

### 4.4. Evaluation of Training Frugality with CLIP

Since an important issue for application purposes is to better understand the possibility of building small training datasets to achieve a reasonable performance, an experiment seeking to check the minimum number of training images per class to obtain acceptable results for fine-tuning CLIP was carried out. The dataset for the analysis was, as mentioned in Section 3.1, a six-class filtered subset of 543 images (434 for training and 109 for testing), with a CLIP model maintaining a learning rate of $10^{-5}$ and a weight decay of $10^{-3}$.

The initial number of train images per class was reduced according to the most limiting class (in this case, we were limited by the *Sarvangasana* class, with 43 training images), which gave us a total of 258 training images. The number of images per class was then halved for each iteration, with the test set of 109 images always kept intact. Results are shown in Table 11. It can be observed that, with 20 training images per class, an accuracy of up to 90% can be achieved.

**Table 11.** Results of the reduction of the number of training images per class experiment.

| Model | Accuracy | Train Images | Test Images | Images by Class |
|---|---|---|---|---|
| | Top-1 | | | |
| ViT-B/32 | 97.2% | 434 | | Unbalanced |
| | 94.5% | 258 | 109 | 43 |
| | **89.9%** | **120** | | **20** |
| | 73.4% | 36 | | 6 |

Best results shown in bold.

## 5. Benchmark with YOLO

The utilization of deep learning models for computer vision tasks has significantly advanced the field, with models such as CLIP and those of the YOLO family. While CLIP excels in versatile visual understanding and classification, YOLOv8 is renowned for its real-time object detection capabilities. The motivation for this comparative analysis stems from the need to evaluate and contrast the efficacy of these two models, each tailored for specific tasks, and to explore potential integration strategies for real-time applications.

YOLOv8 is an advanced model for object detection, renowned for its real-time processing capabilities. Its efficiency lies in partitioning the task of object detection into a single forward pass, allowing for impressive inference speeds while maintaining robust detection performance. This algorithm has become increasingly popular for tasks such as autonomous driving, surveillance, and real-time object tracking, where fast decision-making is essential.

In this section, we compare and contrast fine-tuned CLIP and fine-tuned YOLOv8 in terms of accuracy, fine-tuning cost, and inference time. The accuracy assessment is intended to gauge their performance in the classification task over the Yoga-82 dataset, while the fine-tuning cost will aid in considering the computational resources required to adapt each model to the specific dataset and task. Additionally, the inference time is a pivotal metric, as it quantifies the practicality of integrating these models into real-time systems. The comparison can shed light on their respective strengths and weaknesses and offer insights into their suitability for various real-time visual applications.

Regarding YOLO, after fine-tuning the YOLOv8x-cls model [54] using the Yoga-82 dataset, a maximum accuracy of 87.8% was obtained after 62 epochs. In this benchmark, YOLOv8x-cls and CLIP were both trained and tested using the same sets. The YoPo-subset was then created to compare the accuracy metrics of the two models. The validation set was comprised of a total of 3487 images that were not included in the fine-tuning phase of the models. Figure 8 shows the tuning curve of the YOLOv8x-cls model. Compared to the CLIP fine-tuning curve shown in Figure 5, it took an additional 25 epochs to achieve almost the same level of accuracy. Furthermore, there are indications of accuracy decline in certain epochs throughout the accuracy curve. The fine-tuning time between CLIP and YOLOv8 was also compared, with the latter taking three and a half times longer to reach its maximum accuracy compared to CLIP. On the other hand, where CLIP is clearly disadvantaged is in the inference latency, which is more than 4 times higher than that obtained with YOLO. Figure 9 shows a direct comparison over the validation set between CLIP and YOLO, where the differences between the accuracy of classifying the 82 postures, the fine-tuning cost, and the inference time can be observed. Figure 10 shows the activation map in the same images that were shown by CLIP in Figure 6. These maps were computed using YOLO-V8-CAM [55] to better understand the behavior of YOLOv8 after fine-tuning and to compare the regions on which the YOLOv8 model is focusing when compared to CLIP. We can see that, in the case of the image in the second position, CLIP and YOLOv8 appear to be focusing on the same point after fine-tuning.

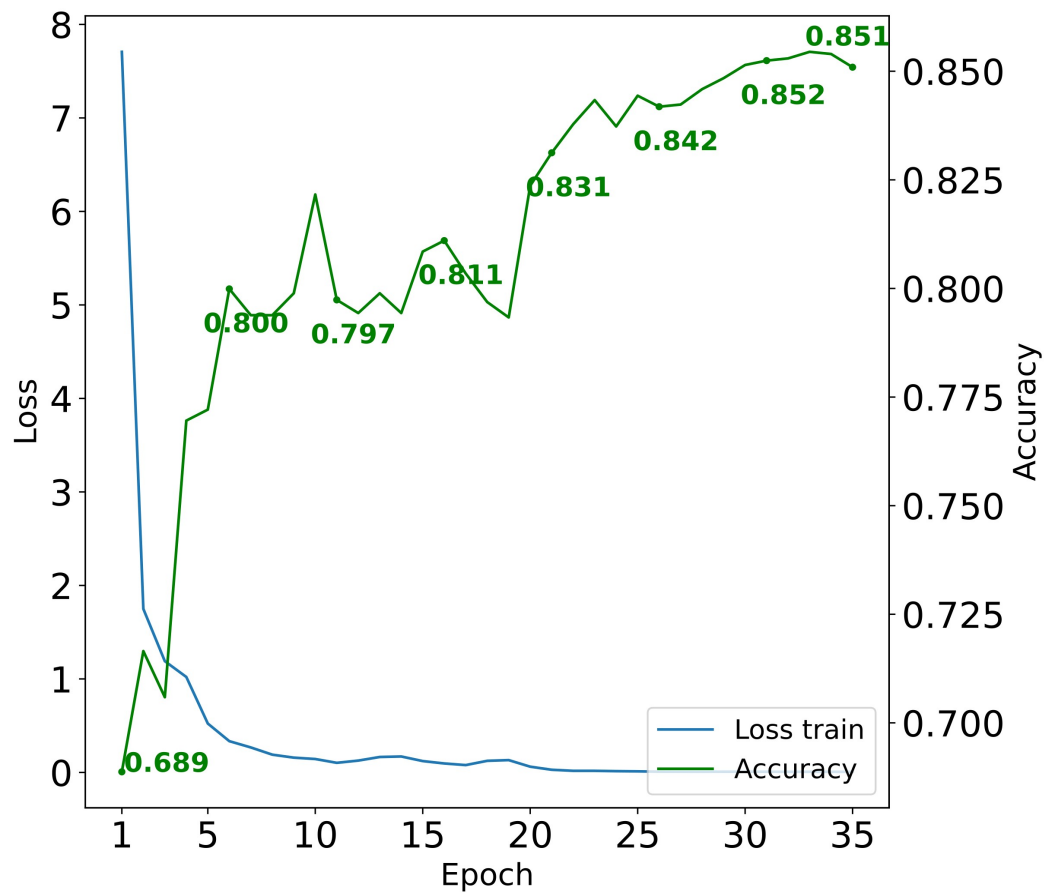

**Figure 8.** Fine-tuning curve of YOLOv8.

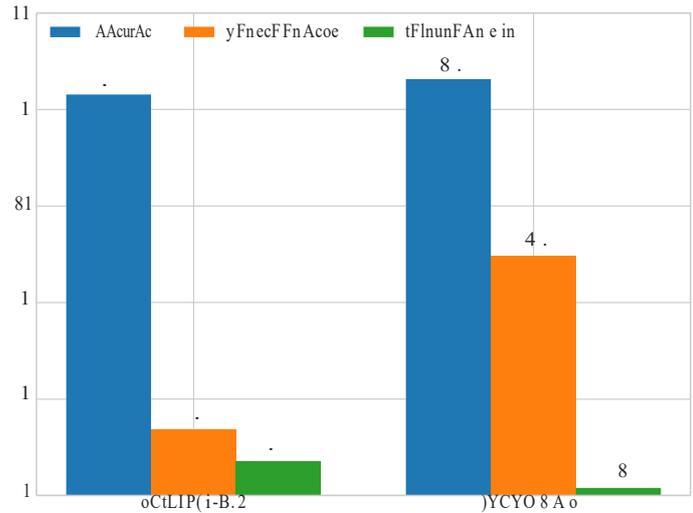

**Figure 9.** Overview. The results of the fine-tuning between CLIP and YOLO are very similar, with almost the same level of accuracy in the ranking of all 82 postures. "Fine-tuning cost" denotes the GPU minutes calculated with a single NVIDIA RTX A4000. "Inference time" is expressed in milliseconds.

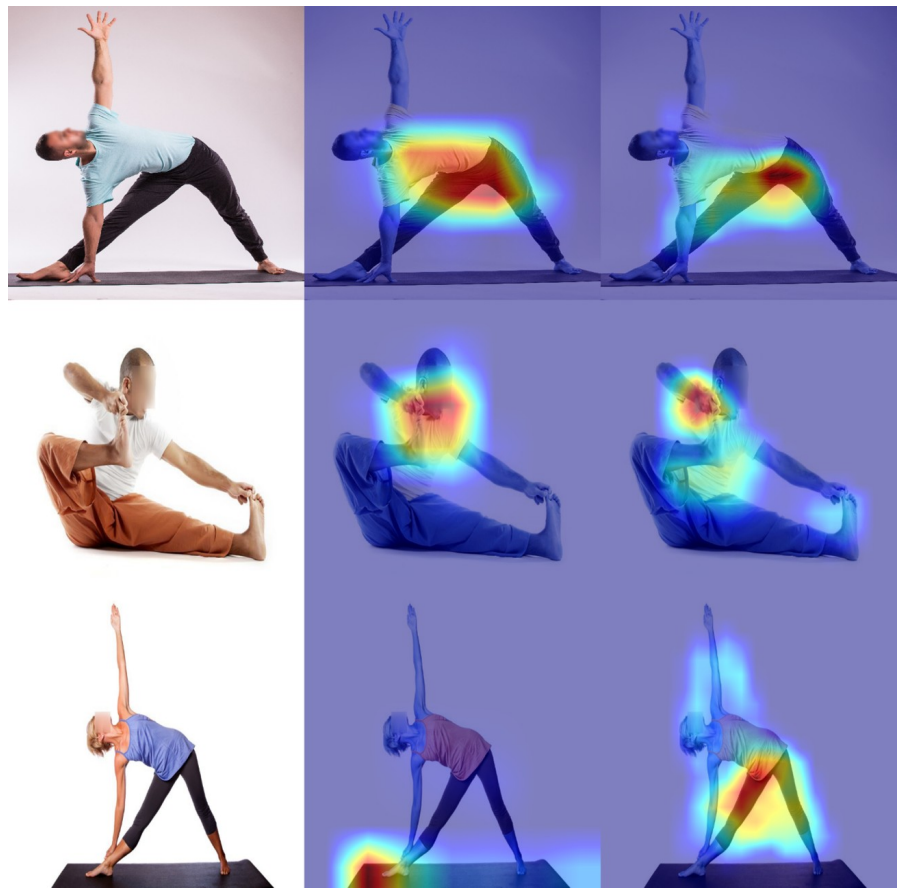

**Figure 10.** Activation map samples with YOLOv8. For each sample shown (**left**), corresponding activation maps from the YOLOv8x-cls pretrained classify model (**middle**) and after fine-tuning (**right**) are shown. The YOLOv8 classification models were pre-trained on the ImageNet dataset with 1000 classes. For the images shown, the predicted classes of the pre-trained model in ImageNet were as follows: Balance Beam with 57% confidence for the top image, Pajama with 47% confidence for the middle image, and Dumbbell with 92% confidence for the bottom image.

## 6. Discussion

Our study demonstrates the potential of CLIP, which has a significant 151 million parameters if considering the ViT-B/32 model, to contribute to the recognition of yoga poses and possibly human postures. The top-1 accuracy of the Yoga-82 dataset in classifying all 82 postures has previously been found to be 79.35% with state-of-the-art models [25]. Through fine-tuning, CLIP achieves an impressive 85.94% accuracy in classifying these 82 yoga poses, which represents a remarkable 6% improvement over previous models. This development is significant to note, although it is crucial to consider that the fine-tuned CLIP model possesses about seven times more parameters than Verma et al.'s work and has over twice the number of parameters in the YOLOv8x-cls pre-trained classification model. The analysis in Section 4.4 also shows that, even with a small dataset of only 20 images per class, fine-tuning CLIP can yield remarkable results, accurately classifying six postures with up to 90% accuracy.

Despite its greater complexity, CLIP sustains competitive training times of only 13.8 min. Comparing it with YOLO, another established model, highlights that CLIP shows a similar accuracy, with a 3.2% difference observed. Both models offer efficient fine-tuning times, but YOLO outperforms CLIP in inference time, requiring only 1.6 ms.

Nevertheless, the findings shown in Section 4.3 hold significant implications for the integration of CLIP into real-time applications, as it provides precise results while maintaining efficiency and enabling faster retraining. One notable aspect that warrants attention is the model's remarkable inference latency, which averaged at 7.1 ms in our experiments. This low latency is instrumental in rendering CLIP highly conducive for real-time applications, where swift decision-making and classification are paramount. The ability to provide accurate and rapid classifications within such a brief timeframe positions CLIP as a viable candidate for a diverse array of real-time systems, ranging from content filtering in social media platforms to the classification and evaluation of body postures.

However, the use of CLIP for posture classification implies certain limitations. In terms of scalability of the model to classify new postures, the desirable zero-shot approach does not provide satisfactory results (as has been shown, few-shot learning is still needed for yoga applications). Moreover, as we explore in Section 3.2.2 (Table 5), the CLIP model uses natural language descriptions to classify images. This need may result in ambiguity or inconsistency in the text descriptions. For example, the same image can be described in different ways, and different images can be described in the same way. This ambiguity can make it difficult for the model to accurately classify images, especially for complex poses or multiclass images (different classes on the same image).

Finally, although we assume that the applied approach can work for other human posture classification tasks (e.g., to support applications in ergonomics or rehabilitation), current research does not allow us to establish a sufficiently strong generalization hypothesis.

## 7. Conclusions

Human pose recognition has long been a formidable challenge in computer vision research due to its broad and diverse applications in everyday life. This article explores the use of a state-of-the-art multimodal learning technique for classification, showing that the fine-tuning of a CLIP model on a complex multiclass dataset can provide reasonable results even with few images per class. CLIP was not designed to serve as a classifier, but as a tool for understanding and interpreting images in the context of natural language, so it has been scarcely explored for classification tasks in previous literature. Our results support that CLIP itself serves as a robust fine-tuner, surpassing the previously established state-of-the-art performance of Verma et al.'s hierarchical yoga pose classification model [25].

There is potential to improve performance by strategically fine-tuning CLIP while freezing certain levels to mitigate model overfitting issues or by using another image (visual) encoder. Future work can include how to generalize the analysis pipeline for human posture classification in other domains, while dealing with rapid adaptation to individual diversity. One strategy to follow is to use generative AI and posterior augmentation to produce individual-tailored datasets. This may also help integrate the model into posture assessment evaluators to, e.g., build personal yoga assistance systems capable of identifying the quality of performance in real time and provide adequate feedback to the user.

**Author Contributions:** Conceptualization, A.M.B.; methodology, A.M.B.; software, A.D.D.; validation, A.D.D.; formal analysis, L.B.; investigation, A.D.D.; data curation, L.B.; writing—original draft, A.D.D., A.M.B. and L.B.; writing—review & editing, A.P. and D.S.-T.; supervision, A.M.B., A.P. and D.S.-T.; funding acquisition, A.M.B. All authors have read and agreed to the published version of the manuscript.

**Funding:** This research was funded by Ministerio de Ciencia e Innovación (PID2020-1034118249RB-C21) and Universidad Politécnica de Madrid (RP220022063).

**Data Availability Statement:** Publicly available datasets were analyzed in this study. This data can be found here: https://sites.google.com/view/yoga-82/home and https://www.kaggle.com/datasets/shrutisaxena/yoga-pose-image-classification-dataset (accessed on 10 November 2023).

**Acknowledgments:** Andrzej Daniel Dobrzycki acknowledges Amazon for the scholarship funding received within the framework of 2022–2023 IPTC UPM-Amazon collaboration programe. IPTC-UPM members acknowledge the support of Roberto Barra-Chicote and the funding received by the European Union under Grant Agreement No. 101103386, by MCIN/AEI/10.13039/501100011033 under grant PID2020-1034118249RB-C21 and by UPM Project RP220022063.

**Conflicts of Interest:** The authors declare no conflict of interest.

## Appendix A. Class Distribution across the Superclasses in the Yoga-82 Dataset

In Section 3.1, the distribution of the 82 postures from the hierarchical level L-3 to the levels L-2 and L-1 is mentioned. The relationship of these postures to the L-2 level and the number of images that could be downloaded from each of them is shown below in Tables A1–A6.

**Table A1.** "Standing" L-1 superclass.

| Standing | | |
|---|---|---|
| Straight | Garudasana | 283 |
| | Vrksasana | 256 |
| | Utkatasana | 312 |
| Forward bend | Parsvottanasana | 183 |
| | Adho Mukha Svanasana | 306 |
| | Ardha Pincha Mayurasana | 69 |
| | Prasarita Padottanasana | 257 |
| | Uttanasana | 390 |
| Side bend | Ardha Chandrasana | 270 |
| | Utthita Trikonasana | 532 |
| | Utthita Parsvakonasana | 530 |
| | Parighasana | 153 |
| | Virabhadrasana I | 172 |
| | Viparita virabhadrasana | 143 |
| | Anjaneyasana | 253 |
| Forward bend | Virabhadrasana II | 259 |
| | Virabhadrasana III | 186 |
| | Natarajasana | 392 |
| | Utthita Padangusthasana | 177 |
| | Urdhva Prasarita Eka Padasana | 133 |

**Table A2.** "Balancing" L-1 superclass.

| Balancing | | |
|---|---|---|
| Front | Bakasana | 294 |
| | Bhujapidasana | 98 |
| | Cockerel | 146 |
| | Tolasana | 125 |
| | Tittibhasana | 184 |
| Side | Parsva Bakasana | 183 |
| | Astavakrasana | 190 |
| | Eka Pada Koundinyanasana I and II | 130 |

**Table A3.** "Reclining" L-1 superclass.

| Reclining | | |
|---|---|---|
| Up-facing | Savasana | 322 |
| | Matsyasana | 288 |
| | Ananda Balasana | 161 |
| | Supta Padangusthasana | 152 |
| | Pawanmuktasana | 156 |
| | Supta Baddha Konasana | 316 |
| | Supta Virasana Vajrasana | 308 |
| | Supta Virasana Vajrasana | 308 |
| | Yogic sleep | 80 |
| Down-facing | Bhujangasana | 788 |
| | Bhekasana | 149 |
| | Salabhasana | 227 |
| | Balasana | 310 |
| | Uttana Shishosana | 82 |
| Side facing | Anantasana | 68 |
| | Vasisthasana | 297 |
| Plank balance | Makara Adho Mukha Svanasana | 46 |
| | Chaturanga Dandasana | 220 |
| | Kumbhakasana | 70 |
| | Mayurasana | 127 |

**Table A4.** "Inverted" L-1 superclass.

| Inverted | | |
|---|---|---|
| Legs straight up | Adho Mukha Vrksasana | 196 |
| | Salamba Sirsasana | 312 |
| | Salamba Sarvangasana | 269 |
| | Pincha Mayurasana | 210 |
| | Viparita Karani | 220 |
| Legs bend | Halasana | 327 |
| | Vrischikasana | 232 |

**Table A5.** "Sitting" L-1 superclass.

| Sitting | | |
|---|---|---|
| Normal 1 | Sitting | 671 |
| | Baddha Konasana | 268 |
| | Malasana | 241 |
| | Dandasana | 83 |
| | Pasasana | 38 |
| Normal 2 | Gomukhasana | 303 |
| | Vajrasana | 329 |
| | Bharadvajasana I | 83 |
| | Ardha Matsyendrasana | 312 |
| Split | Split | 302 |
| | Upavistha Konasana | 188 |
| Forward bend | Janu Sirsasana | 180 |
| | Parivrtta Janu Sirsasana | 220 |
| | Paschimottanasana | 348 |
| | Tortoise | 149 |
| Twist | Akarna Dhanurasana | 84 |
| | Krounchasana | 58 |
| | Rajakapotasana | 274 |

**Table A6.** "Wheel" L-1 superclass.

| Wheel | | |
|---|---|---|
| Up-facing | Urdhva Dhanurasana | 308 |
| | Dwi Pada Viparita Dandasana | 72 |
| | Purvottanasana | 225 |
| | Kapotasana | 53 |
| | Setu Bandha Sarvangasana | 279 |
| | Camatkarasana | 178 |
| | Ustrasana | 311 |
| Down-facing | Marjaryasana | 409 |
| Others | Dhanurasana | 228 |
| | Paripurna Navasana | 394 |

## Appendix B. Confusion Matrices

The confusion matrices of the L-1 "Reclining" superclass in Figure A1 and the "Sitting" superclass in Figure A2 are shown below.

**Table A7.** "Reclining" L-1 superclass label-name correspondence.

| "Reclining" L-1 Superclass Label-Name Correspondence | | | |
|---|---|---|---|
| 0 | Savasana | 1 | Matsyasana |
| 2 | Ananda Balasana | 3 | Supta Padangusthasana |
| 4 | Pawanmuktasana | 5 | Supta Baddha Konasana |
| 6 | Supta Virasana Vajrasana | 7 | Yogic sleep |
| 8 | Bhujangasana | 9 | Bhekasana |
| 10 | Salabhasana | 11 | Balasana |
| 12 | Uttana Shishosana | 13 | Anantasana |
| 14 | Vasisthasana | 15 | Makara Adho Mukha Svanasana |
| 16 | Chaturanga Dandasana | 17 | Kumbhakasana |
| 18 | Mayurasana | | |

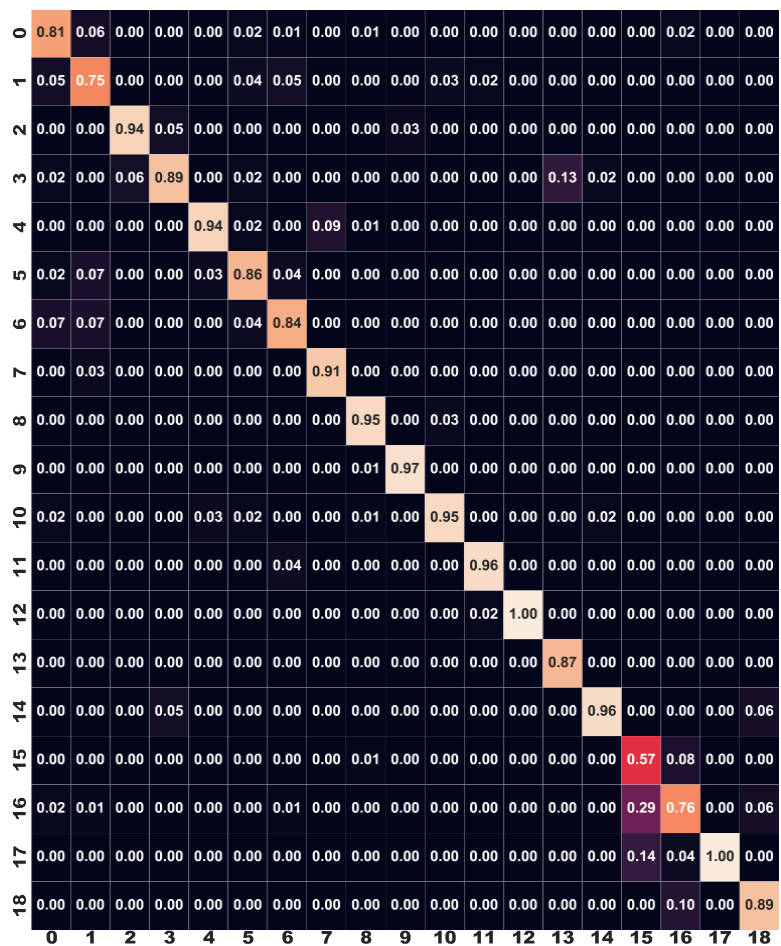

**Figure A1.** Hierarchical confusion matrices. "Reclining " L-1 superclass. It can be observed that CLIP has problems classifying the postures Makara Adho Mukha Svanasana and Chaturanga Dandasana. Lighter color values denote better performance.

**Table A8.** "Sitting" L-1 superclass label-name correspondence.

| "Sitting" L-1 Superclass Label-NAME Correspondence | | | |
|---|---|---|---|
| 0 | Sitting | 1 | Baddha Konasana |
| 2 | Malasana | 3 | Dandasana |
| 4 | Pasasana | 5 | Gomukhasana |
| 6 | Vajrasana | 7 | Bharadvajasana I |
| 8 | Ardha Matsyendrasana | 9 | Split |
| 10 | Upavistha Konasana | 11 | Janu Sirsasana |
| 12 | Parivrtta Janu Sirsasana | 13 | Paschimottanasana |
| 14 | Tortoise | 15 | Akarna Dhanurasana |
| 16 | Krounchasana | 17 | Rajakapotasana |

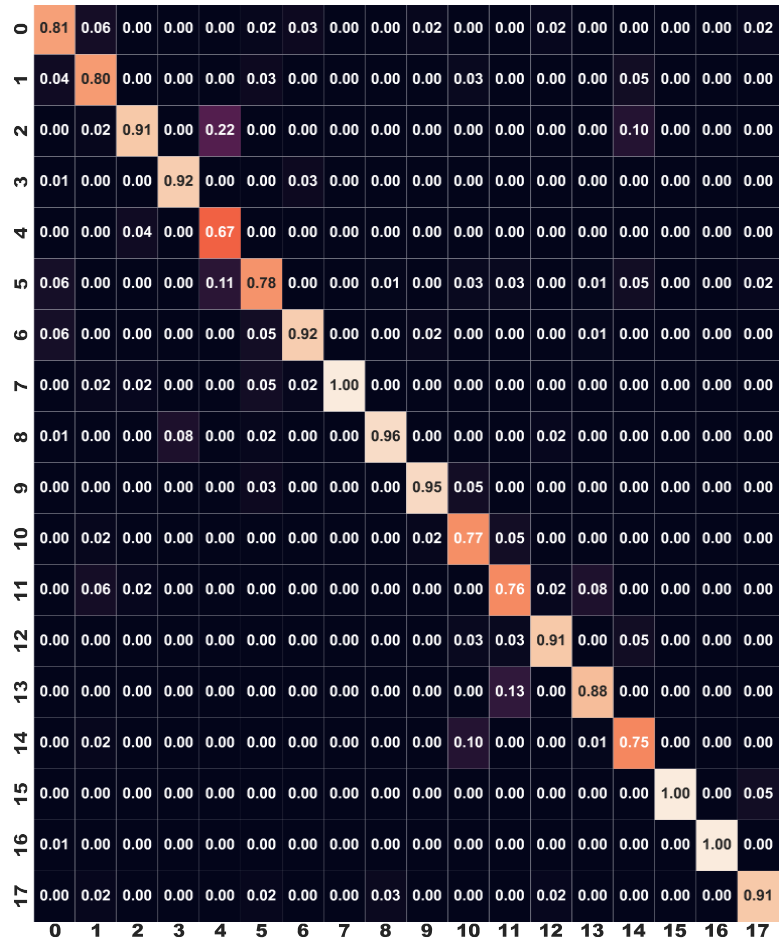

**Figure A2.** Hierarchical confusion matrices. "Sitting" L-1 superclass. Lighter color values denote better performance.